\documentclass[runningheads]{llncs}
\usepackage[T1]{fontenc}
\usepackage{graphicx}
\usepackage{amsmath}
\usepackage{hyperref}

\begin{document}
\title{Temporal-Difference Learning for Dragonchess}

\author{Jim O'Connor \and
Annika Hoag \and
Sarah Goyette \and
Gary B. Parker}
\authorrunning{J. O'Connor et al.}
% First names are abbreviated in the running head.
% If there are more than two authors, 'et al.' is used.
%
\institute{Connecticut College, New London CT 06320, USA \\
\email{\{joconno2, ahoag, sgoyette, parker\}@conncoll.edu}}

\maketitle              % typeset the header of the contribution

\begin{abstract}
Our research investigates how two adaptive AI methods, evolutionary transfer learning and TD($\lambda$), perform in the three-dimensional chess environment Dragonchess. The game challenges players with its unique board structure and computational load, making it an ideal setting to study how adaptive methods can update evaluation heuristics in novel environments. In this work we re-implement the Dragonchess engine, changing it from a PyGame engine to C++. This enables faster gameplay, allowing us to run 10,000 games with confidence intervals and significance tests, rather than a single small tournament. Both adaptive methods outperform all other agents in the round-robin tournament. Our results showed that there is no significant difference in the performance between the evolved and learned evaluations. This research establishes the efficacy of adaptive methods in structurally complex, novel game domains.

\keywords{temporal-difference learning \and CMA-ES \and decisive games}
%took from glossary in GitHub so may need to be edited/some of these are the same as before 
\end{abstract}

\section{Introduction}
Game playing has demonstrated to be a productive testing ground for developments in the domain of artificial intelligence (AI). From early heuristic-based strategies in chess and checkers \cite{chole2024evolution} to modern deep reinforcement learning approaches exemplified by systems like AlphaZero \cite{silver2017mastering}, the field has progressed substantially in tailoring AI to intricate strategic settings. However, much of the previous research focused on games with moderately static board structures, primarily two-dimensional layouts with well-defined, stable piece-movement rules \cite{hu2024games}. As AI strategies continue to advance, there is an ability to extend these techniques to increasingly complex game environments that present novel structural and strategic challenges.

Dragonchess, a game developed by Gary Gygax, offers an interesting case for studying AI adaptation in novel game environments. In contrast to traditional chess, Dragonchess features a three-tiered board, substantially increasing the state and action space of the game \cite{DragonChess}. Its unique framework offers multilevel interactions, distinct movement rules for different piece types, and challenges players to be more forward-thinking than conventional chess-like games. These factors make Dragonchess an immensely challenging domain for game-playing research, as more traditional search-based strategies and heuristic evaluations struggle with the domain's increased complexity. 

Our previous work investigated evolutionary transfer learning in the context of Dragonchess by adapting heuristics from Stockfish, the leading open-source chess engine \cite{stockfish2025}. While these heuristics are optimized for conventional chess, they perform poorly when directly applied to Dragonchess, prompting us to employ the Covariance Matrix Adaptation Evolution Strategy (CMA-ES) \cite{ostermeier1994derandomized} to evolve them. This work showed that through the application of CMA-ES, transferred Stockfish heuristics can adapt to a competitive Dragonchess evaluation function, resulting in an agent that outperformed handcrafted Jackman and Gygax weightings in a Swiss tournament \cite{oconnorevoltransfer}. This motivated us to explore how a learning-based method would compare to the evolutionary agent. 

In this extension, we create an agent that incorporates a temporal-difference self-play TD($\lambda$). TD($\lambda$) enables the agent to improve its evaluation function without external supervision or search labels \cite{TDlamb}. Transfer learning and TD($\lambda$) differ significantly. The CMA-ES evolved agent scales a fixed set of transferred heuristics using a $14$-dimensional piece-value vector. In contrast, the learning-based agent fits a $40$-feature linear evaluation directly from self-play. It combines per-piece material terms with Dragonchess-specific positional signals such as frozen-piece counts, pieces stationed above certain enemies, and king-proximity counts. To compare these agents, we re-engineered the Dragonchess platform, replacing the original PyGame engine with an open-source C++ engine. It includes alpha-beta search and a transposition table, is more efficient, running thousands of decisive games per matchup, and reports statistical confidence intervals rather than a single tournament ranking.

Our tests show that both adaptive methods, evolutionary and learning, surpass the handcrafted baseline and plain search, validating the effectiveness of adaptive evaluation in Dragonchess. We illustrate that these policies are viable strategies for AI in intricate and underrepresented game domains. 

On a wider scale, this work demonstrates that multiple adaptive AI-driven strategies can be applied to a three-dimensional board game, highlighting the importance of heuristic adaptation in novel domains. By illustrating the effectiveness of adaptive evaluation, whether through evolution or self-play learning, this research furthers the study of generalizable AI approaches to strategic decision-making.

\section{Related Work}

Research on artificial intelligence in game-playing contexts has primarily focused on deterministic, fully observable domains such as chess, Go, and shogi. In these settings, deep search methods and heuristic evaluation functions have driven progress in computational play. Early chess engines, including Deep Blue \cite{campbell2002deep}, illustrated the effectiveness of brute-force search when combined with refined heuristics. Subsequent approaches, such as Monte-Carlo Tree Search (MCTS), introduced simulation-based search, which proved extremely effective in games like Go \cite{gelly2012grand}. Recent advances in deep reinforcement learning have emphasized self-play and learned policy networks, as illustrated by AlphaZero \cite{silver2017mastering} and MuZero\cite{schrittwieser2020mastering}. Although these methods have achieved notable success, they require substantial training time and computational resources, limiting their practicality for various emerging domains, like games with complex three-dimensional structures such as Dragonchess \cite{silver2017mastering3}.

Temporal Difference learning was formally introduced by Sutton as a method of using past experiences in an unknown environment to predict future outcomes in multi-step prediction problems \cite{TDorginal}. Building on this primary work, research expanded into a wide range of fields. This includes game playing, in which agents learn effective policies through self-play \cite{TDchess}. Temporal-difference learning has also been used in finance to solve optimal execution problems \cite{TDfin}. Another strategy is TD($\lambda$), enabling temporal-difference methods to solve more complex problems. TD($\lambda$) has a computational cost similar to traditional TD methods but learns at a faster rate \cite{TDlamb}. 

Adaptive methods, such as temporal-difference learning and transfer learning, have proven effective in game-playing. Tesauro created a state-of-the-art backgammon agent utilizing TD($\lambda$) \cite{TDback}. Evolutionary transfer learning has also shown promise in optimizing AI agents for complex strategic tasks by evolving and fine-tuning existing heuristics \cite{hou2019evolutionary}.

\section{Dragonchess}
Dragonchess, originally created in 1985 by Gary Gygax in Dragon magazine, is a three-dimensional chess variant that introduces additional strategic complexity beyond traditional chess \cite{DragonChess}. The game consists of three vertically stacked 12×8 boards, each portraying various terrain types: the top board represents the sky, the middle board symbolizes the land, and the bottom board corresponds to the underworld. Each board offers unique interactions and move restrictions, challenging players to think beyond the two-dimensional constraints of conventional chess.

\begin{figure}[!ht]
  \centering
  \includegraphics[width=0.7\textwidth]{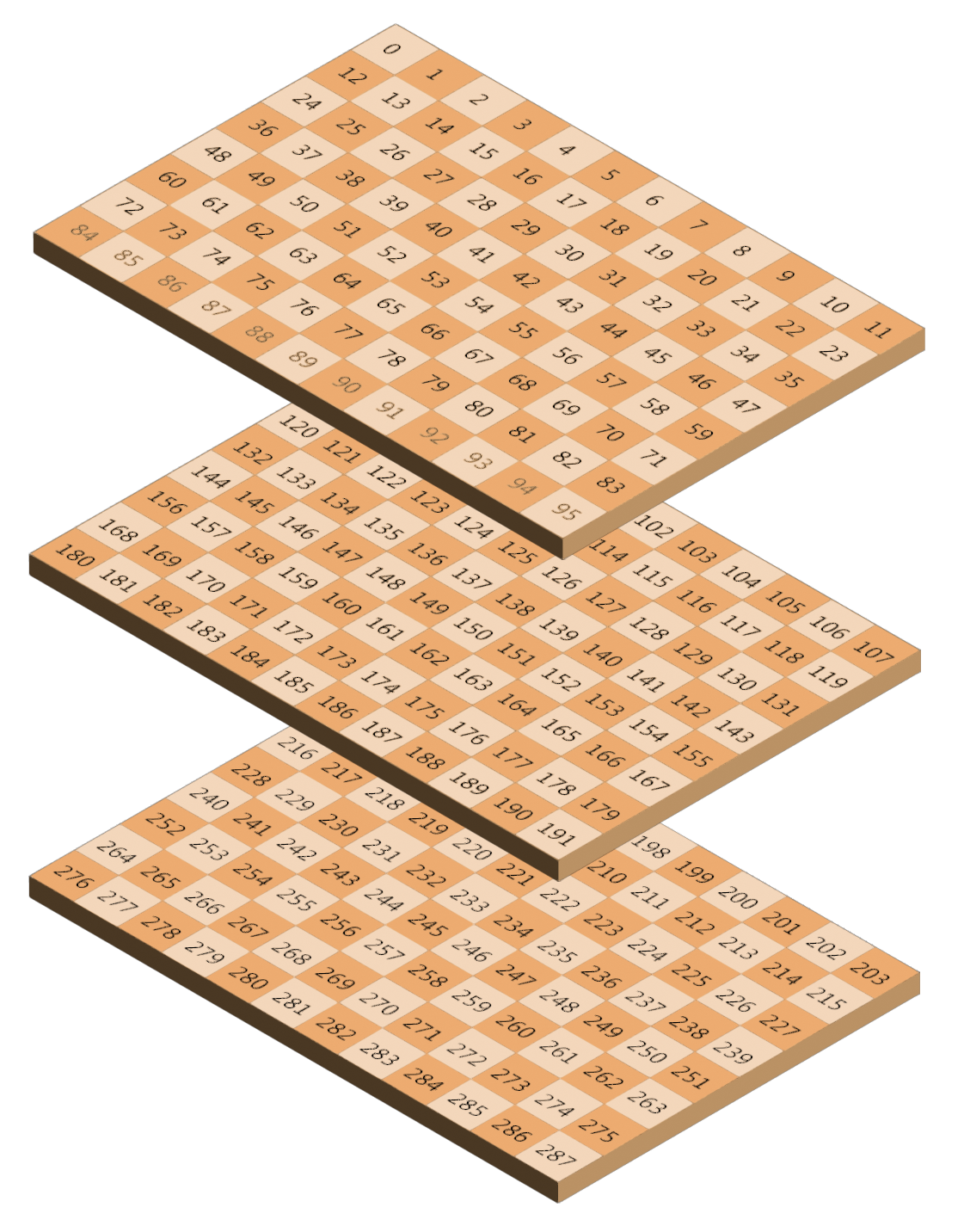}
  \caption{Each layer of the board (Sky, Ground, and Underworld) consists of 8 rows and 12 columns, represented by integer indices from 0 to 287. At any given index of the array, an integer constant is stored, representing piece type and ownership. These constants are positive for Gold, and negative for Scarlet. This is the same structure as the original Dragonchess implementation \cite{oconnorevoltransfer}.}
  \label{fig1}
 \end{figure}

Dragonchess offers varied gameplay, introducing inter-board mobility and atypical movement patterns. For instance, the Basilisk, occupying the "Underworld" board, can briefly freeze opponents on the "Land" board, adding a unique strategic dimension. The Dragon, an extremely influential piece, remains on the top board but has the unparalleled ability to capture pieces on the middle board without moving. Other pieces, such as the Griffin and the Hero, can move vertically between boards. These complex interactions highlight the significance of vertical placement and inter-board strategy. 

One of Dragonchess' most challenging features is its considerably expanded state space. Standard chess offers roughly $10^{43}$ legal configurations \cite{Shannon}. Dragonchess' three-tiered board structure dramatically increases this number to around $10^{63}$. The game's unique movement system introduces novel strategic decisions, such as restricting enemy movement by controlling vertical positions or organizing multi-board assaults. These elements of the game make heuristic evaluation challenging, as more traditional evaluation functions struggle to capture the intricate interplay between boards. 

In Dragonchess, AI techniques such as traditional search struggle with the game's immense action space, since interactions occur across three boards rather than on a 2D plane. Previously established chess heuristics, such as board control and piece mobility, do not directly translate to Dragonchess, as piece values vary significantly with board position and potential vertical mobility. Some pieces, like a Mage, are very mobile on the Land board but are limited on the Sky or Underworld boards, requiring heuristic adjustment to account for these transitions. 

These complex systems make Dragonchess a productive testbed for evaluating heuristic adaptation and evolutionary optimization techniques. The difficulties traditional chess heuristics encounter in Dragonchess suggest investigating adaptive optimization methods so the weight of these metrics can be adapted to a multi-board environment. Understanding these challenges aids in creating stronger Dragonchess-playing AI and offers broader insights into the adaptability of heuristic-based AI in novel strategic environments.

\section{Methodology} 
Our previous work demonstrated that CMA-ES \cite{ostermeier1994derandomized} can be used to adapt transferred Stockfish heuristics into an evaluation function for the Dragonchess chess variant. This system outperformed the handcrafted Jackman and Gygax weights in a Swiss style tournament \cite{oconnorevoltransfer}. Extending this work, we add temporal-difference self-play (TD($\lambda$)). We compare TD($\lambda$) with the evolved evaluator under controlled conditions, as detailed later in this section.

We also introduce a re-engineered experimental platform, replacing the original PyGame engine \cite{oconnorevoltransfer} with an open-source C++ engine. Our new platform has alpha-beta search and a transposition table, which is fast enough to run thousands of decisive games per matchup. The C++ engine can also report statistical confidence intervals as opposed to a single tournament ranking.

\subsection{Learned Evaluation Using Self-Play TD($\lambda$)}
The previous evolutionary transfer learning approach resulted in an agent which scales a fixed set of transferred Stockfish heuristics with a 14-dimensional piece-value vector \cite{oconnorevoltransfer}. In this extended version, the learned agent fits a 40-feature linear evaluation directly from self-play. These features combine per piece material terms with Dragonchess specific positional signals, such as frozen-piece counts and material, pieces stationed above the enemy's Basilisk, and king-proximity counts. Raw feature values are used directly with no feature normalization or regularization.

Weights are trained by TD($\lambda$) ($\lambda{=}0.7$, $\gamma{=}1.0$, learning rate $0.05$) with a linear value function $V(s)=w^\top\phi(s)$, accumulating eligibility traces, and batch-accumulated gradient ascent. Per ply, the temporal difference error is calculated as $\delta_t=\gamma V(s_{t_1})-V(s_t)$. Meanwhile, terminal states use $\delta_T=z-V(s_T)$, where the terminal reward $z$ is encoded from Gold's perspective as $+1$ for a Gold win, $-1$ for a Scarlet win, and $0$ for a draw. Eligibility traces are updated as $e\leftarrow\gamma\lambda e+\phi(s+t)$, reset at the start of each game. The accumulated gradient is summed across batches of 50 self-play games before a single AdaGrad update is applied. This system requires no external supervision of search labels, the signal is the agent's own outcome.

During training, exploration is limited, there is no $\epsilon$-greedy, no temperature, and no randomized openings. The only stochasticity is random tie-breaking among nearly equivalent moves and a mixed self-play schedule. This schedule entails $70\%$ of batches using the current weights and the remaining $30\%$ playing against a frozen snapshot of a previous policy. Draws receive a neutral reward of $0$ and area declared after a stalemate, 100 plies after capture, or reaching the 1000-move limit.

Training ran for 22,000 games. Every fixed number of batches, the current weights were evaluated against a depth-2 alpha-beta opponent on a held-out set, with the best-performing checkpoint retained. During training, this checkpoint reached a 0.67 win rate against a depth-2 alpha-beta opponent.

\subsection{Depth-Matched Tournament Protocol}
In order to separate evaluation quality from search effort, every agent uses the same alpha-beta depth of two plies. The only variable is the evaluation function. We compare five agents:
\begin{itemize}
    \item \textbf{CMA-ES}: Our evolved evaluator using evolutionary transfer learning \cite{oconnorevoltransfer}
    \item \textbf{TD($\lambda$) Self-Play}: Learned temporal-difference self-play evaluator (\emph{TD($\lambda$)})
    \item \textbf{Simple Alpha-Beta}: Uses the engine's default material evaluation 
    \item \textbf{Jackman}: Basic alpha-beta using piece weights recommended by Edward Jackman specifically for Dragonchess
    \item \textbf{Random Agent}: Baseline
\end{itemize}
We omit the Gygax baseline used in our previous Dragonchess work because Gygax's original article gives only qualitative relative strength statements without any numeric piece values. Because of this, any vector would be fabricated. Jackman, on the other hand, is a stronger handcrafted reference and is a harder bar to clear.

Each unordered pair of agents plays 1000 games split evenly across colors (i.e., 500 games with each agent getting to move first). This removes a first-move bias which the previously used single-pass Swiss-style tournament did not account for. Our full study contains 10000 games, two orders of magnitude more than our previous work's single 50-round Swiss \cite{oconnorevoltransfer}. The new C++ engine completes these games in roughly six minutes. Ratings are computed with Bradley-Terry maximum likelihood (draws scored as one half), with the field mean anchored to 1500 and a scale of 400. The significance of each head-to-head competition is assessed on decisive games only with a two-sided binomial test and 95\% Wilson  confidence intervals.

\subsection{C++ Dragonchess Game Engine}
The study done in this work was enabled by our re-implementation of the Dragonchess game engine into C++. Our initial work utilized a PyGame engine, evaluating at three seconds per move. Therefore, a single game of a few hundred plies cost on the order of minutes, which bounded our empirical study to a single small tournament \cite{oconnorevoltransfer}.

\begin{figure*}[!ht]
  \centering
  \includegraphics[width=\linewidth]{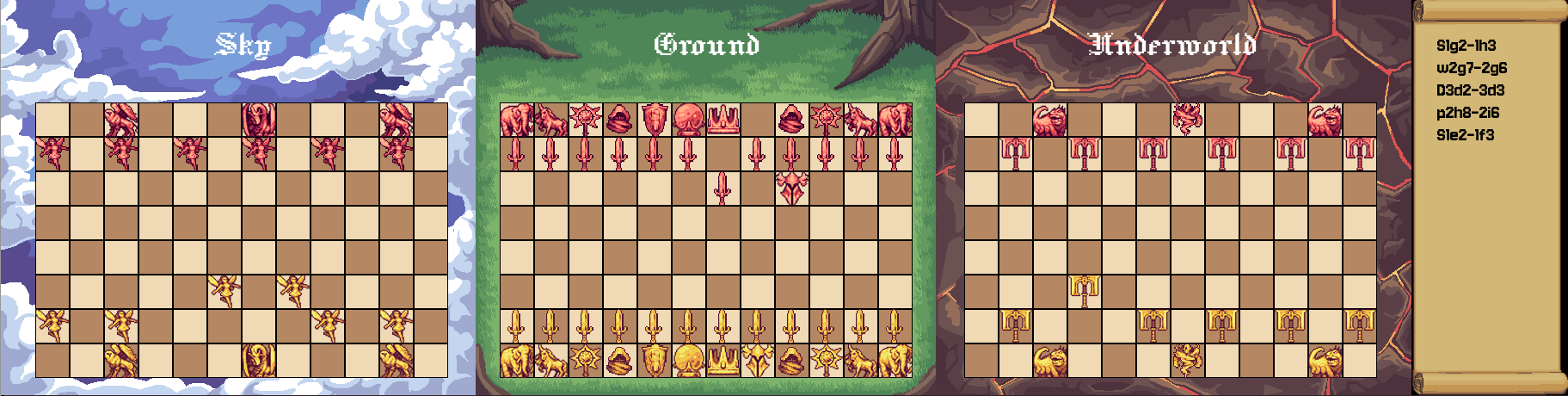}
  \caption{Screenshot from our Dragonchess engine's graphical interface built using C++. Dragonchess' characteristic three-layered board can be seen with the Sky (top), Land (middle), and Underworld (bottom)—with distinct piece types and clear depiction of multi-layer interactions and movements unique to Dragonchess. This is the same GUI as in the original PyGame implementation \cite{oconnorevoltransfer}.}
  \label{fig2}
 \end{figure*}
 
The new C++ engine uses alpha-beta search with a transposition table and a typed move generator. With the depth-2 setting used in this study, it plays a complete self-play game (averaging several hundred plies) in approximately 85 ms. The engine sustains around 12 games per second on 16 threads for the draw-heavy, strong agent matches, and even faster for more decisive ones. The full 10,000-game round-robin finishes in about six minutes on a single workstation, making the confidence intervals and significance tests affordable. Previously, our PyGame engine could report one ranking, whereas the same hardware budget here yields hundreds of decisive games per pairing. The engine is open-source, released alongside the original Python implementation\footnote{\url{https://github.com/joconno2/DragonchessAI-Engine}}. Figure ~\ref{fig2} shows the GUI of our game engine.

\section{Results}
\begin{figure}[!ht]
    \centering
    \includegraphics[width=1.2\textwidth]{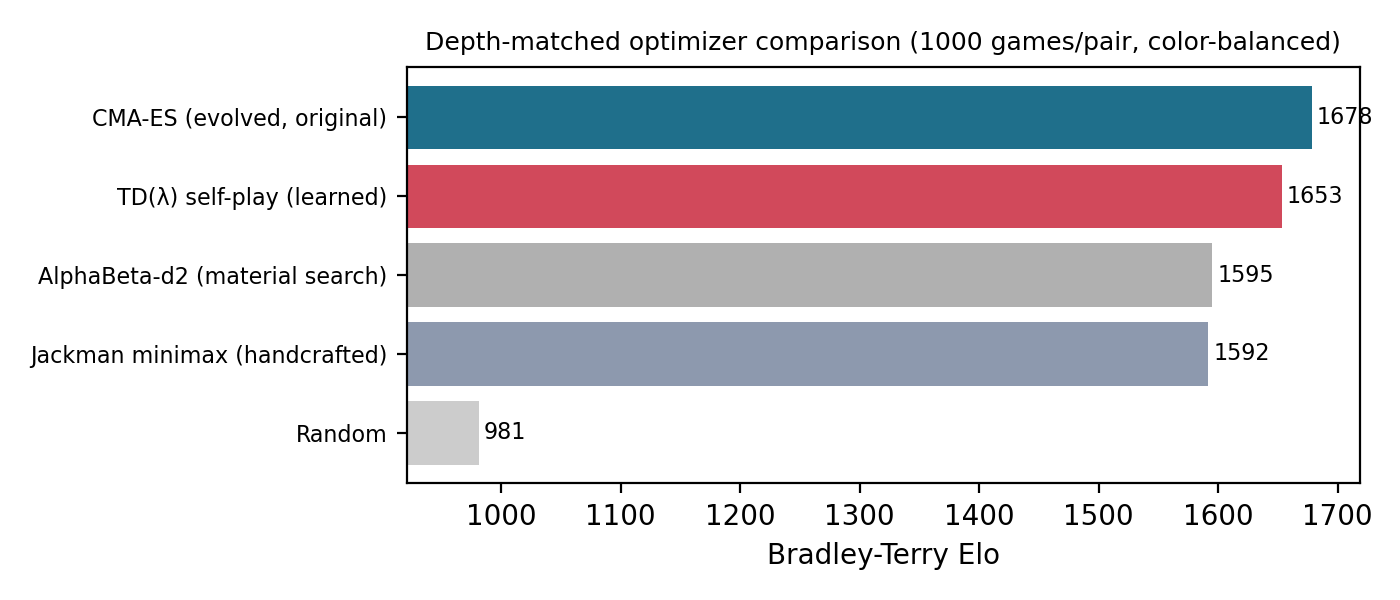}
    \caption{Bar chart visualization of the Bradley-Terry Elo round-robin standings detailed in Table~\ref{tab:standings}.}
    \label{fig3}
 \end{figure}

Table~\ref{tab:standings} and Figure~\ref{fig3} show the final standings of each of our five agents; Table~\ref{tab:pairwise} details the head-to-head matches. Two findings stand out: our adaptive evaluation methods (bolded in Table~\ref{tab:standings}) were the strongest performers, and the evolved and learned evaluators are indistinguishable when going head-to-head, statistically speaking.

\begin{table}[h]
\centering
\caption{Depth-matched (AlphaBeta $d{=}2$) round-robin standings, 1000 games per pair, color-balanced. Elo by Bradley-Terry MLE (draws as 0.5), mean anchored to 1500.}
\label{tab:standings}
\begin{tabular}{lrrrrr}
\hline
Agent & W & L & D & Score & Elo \\
\hline
CMA-ES (evolved, original) & \textbf{2131} & \textbf{615} & \textbf{1254} & \textbf{0.690 }& \textbf{1678.4} \\
TD($\lambda$) self-play (learned) & \textbf{2314} & \textbf{1088} & \textbf{598} & \textbf{0.653} & \textbf{1653.1 }\\
AlphaBeta-$d2$ (material search) & 1932 & 1380 & 688 & 0.569 & 1595.2 \\
Jackman (handcrafted) & 1889 & 1374 & 737 & 0.564 & 1592.0 \\
Random & 47 & 3856 & 97 & 0.024 & 981.3 \\
\hline
\end{tabular}
\end{table}

\begin{table}[!h]
\centering
\caption{Head-to-head decisive win rates (draws excluded) with 95\% Wilson confidence intervals. Search depth fixed at 2.}
\label{tab:pairwise}
\begin{tabular}{llrrr}
\hline
A & B & A rate & 95\% CI \\
\hline
TD($\lambda$) & CMA-ES & 0.503 & [0.463, 0.544] \\
TD($\lambda$) & Jackman & 0.601$^{*}$ & [0.569, 0.632]  \\
TD($\lambda$) & AlphaBeta & 0.560$^{*}$ & [0.528, 0.591]  \\
TD($\lambda$) & Random & 0.982$^{*}$ & [0.972, 0.989] \\
CMA-ES & Jackman & 0.743$^{*}$ & [0.706, 0.777] \\
CMA-ES & AlphaBeta & 0.758$^{*}$ & [0.723, 0.790]  \\
CMA-ES & Random & 0.976$^{*}$ & [0.964, 0.984]  \\
Jackman & AlphaBeta & 0.497 & [0.461, 0.532]  \\
Jackman & Random & 0.997$^{*}$ & [0.991, 0.999]  \\
AlphaBeta & Random & 0.996$^{*}$ & [0.990, 0.998]  \\
\hline
\end{tabular}
\end{table}

Both adaptive methods clearly outperform both the handcrafted baseline and plain search, as demonstrated in Table~\ref{tab:standings}. The evolved evaluator beats Jackman on $0.743$ of decisive games ($p<10^{-32}$) and alpha-beta on $0.758$ ($p<10^{-39}$), while the learned evaluator beats Jackman on $0.601$ ($p<10^{-9}$) and alpha-beta on $0.560$ ($p<10^{-3}$). The handcrafted Jackman agent is statistically indistinguishable from simple depth-2 search ($0.497$, $p=0.88$), confirming that the gains we see in the adaptive methods are coming from adapted evaluation, not from search. This bolsters the claim from our previous work: gradient-free adaptation of a transferred evaluation outperforms handcrafted weights \cite{oconnorevoltransfer}. These results also demonstrate how this holds for a second, mechanistically different method.

The second finding we can see in Table~\ref{tab:pairwise}: the evolved and learned evaluators are not statistically different in head-to-head matches. Over 1000 games, TD($\lambda$) scores $0.503$ of decisive games against CMA-ES (95\% CI $[0.463, 0.544]$, $n{=}576$ decisive). Further, the CMA-ES agent has the top overall Elo ($1678$ vs.\ $1653$) by defeating the weaker baselines more decisively and with a higher draw rate, not by beating the learned agent. Therefore, a self-play learner with no search supervision reaches the same performance tier as the evolved evaluator which required a population-based outer loop.

\section{Conclusion}
The study completed in this paper highlights the generality of the work presented in \cite{oconnorevoltransfer}, demonstrating that adapting a transferred evaluation by either evolution or self-play learning, yields agents that are in the same strength tier and significantly above handcrafted weights. These results show that the learned evaluator matches (not surpasses) the evolved one at the tested search depth. 

Our study has two limitations that leave room for future investigation. First, draw rates tend to be high among the strongest agents, up to $\sim40\%$. This is a property of shallow and material-dominated play. Second, the learned and evolved agents utilize different feature parameterizations, 40 learned features vs. 14 evolved piece values. Therefore, the comparison is between methods with their natural representations, as opposed to a single representation optimized two ways. A representation-matched study, where the 40-feature vector is evolved with CMA-ES or warm-starting evolution from the learned weights, is a natural future direction.

%
% ---- Bibliography ----
%
% BibTeX users should specify bibliography style 'splncs04'.
% References will then be sorted and formatted in the correct style.
%
\bibliographystyle{splncs04}
\bibliography{references}
\end{document}